\documentclass{article}
\usepackage{spconf,amsmath,epsfig}

\usepackage{algorithm}
\usepackage{algorithmic}
\usepackage{enumitem}

\usepackage[colorlinks,
			linkcolor=blue,
			anchorcolor=blue,
			citecolor=green]{hyperref}
\usepackage{amsmath}
\usepackage[scr=boondox, cal=esstix]{mathalpha}
\usepackage{cuted}
\usepackage{amssymb}
\usepackage{booktabs}
\usepackage{subfigure}

\let\OLDthebibliography\thebibliography
\renewcommand\thebibliography[1]{
  \OLDthebibliography{#1}
  \setlength{\parskip}{0pt}
  \setlength{\itemsep}{0pt plus 0.3ex}
}

\begin{document}\sloppy


\title{Out-of-Distribution Detection Using Peer-Class Generated by Large Language Model}
%
\name{K huang, G song, hanwen Su, jiyan Wang}
\address{}

\maketitle

\begin{abstract}
Out-of-distribution (OOD) detection is a critical task to ensure the reliability and security of machine learning models deployed in real-world applications. Conventional methods for OOD detection that rely on single-modal information, often struggle to capture the rich variety of OOD instances. The primary difficulty in OOD detection arises when an input image has numerous similarities to a particular class in the in-distribution (ID) dataset, e.g., wolf to dog, causing the model to misclassify it. Nevertheless, it may be easy to distinguish these classes in the semantic domain. To this end, in this paper, a novel method called ODPC is proposed, in which specific prompts to generate OOD peer classes of ID semantics are designed by a large language model as an auxiliary modality to facilitate detection. Moreover, a contrastive loss based on OOD peer classes is devised to learn compact representations of ID classes and improve the clarity of boundaries between different classes. The extensive experiments on five benchmark datasets show that the method we propose can yield state-of-the-art results.
\end{abstract}
\section{Introduction}
\label{sec:intro}

Out-of-distribution detection is an essential endeavour in machine learning and artificial intelligence systems, especially in applications where the models need to make reliable predictions in open-world environments. OOD instances refer to data samples that deviate significantly from the distribution of the training data. This often leads to unreliable model predictions or even catastrophic errors, e.g., when a medical image is interpreted as a positive sample due to an unknown abnormality in a known area\cite{GaoW20}. Therefore, the accurate identification of OOD data as ”unknown” is of utmost importance for the robustness and security of machine learning systems.


There are numerous methods for OOD detection, including confidence-based methods \cite{Esmaeilpour2021ZeroShotOD},\cite{fort2021exploring}, density-based methods \cite{NEURIPS2020_eddea82a},\cite{NEURIPS2021_7d3e28d1}, and distance-based methods \cite{sun2022knnood},\cite{vanamersfoort2020uncertainty}. Confidence-based methods use the score function, while density-based methods use the reconstruction loss in generative models to determine whether a sample is an OOD sample. Distance-based methods use the feature embeddings extracted from a model and assume that the test OOD samples are relatively far away from the ID data. Most existing methods have focused primarily on the use of single-modal information to distinguish between in-distribution and OOD samples. However, they are restricted to capturing limited information from a specific modality, e.g., image, rather than collecting comprehensive descriptions from multi-modality to represent the OOD data for learning detection.

Recently, large-scale models extended by transformers \cite{NIPS2017_3f5ee243} have gained popularity in various fields of research. Large Language Models (\textbf{LLMs}) show an impressive ability to perform semantic tasks with zero-shot learning. This zero-shot ability underlines their exceptional semantic understanding and their ability to adapt to different linguistic tasks through latent knowledge representation. It exhibits the ability to handle novel scenarios, understand context, and generate coherent text even in unfamiliar contexts, i.e., OOD semantics. This insensitivity to distribution shifts makes them versatile tools for various NLP tasks and contributes to their reliability in real-world applications. In addition, Vision Language Pre-trained models (\textbf{VLMs}) offer a transformative approach to understanding and processing vision and language data. The architectural framework of \textbf{VLMs} represents a harmonious combination of transformer-based text and image encoders. This design makes it possible to effectively bridge the gap between textual and visual domains, enabling a more comprehensive and contextually rich understanding of vision data. CLIP \cite{pmlr-v139-radford21a} is a state-of-the-art multi-modal pre-trained model that achieves remarkable performance on image-text tasks by training jointly on a diverse image and text corpus and effectively learning to associate textual and visual content to achieve robust cross-modal comprehension. Therefore, it is possible to utilize the ability of \textbf{LLMs} to handle OOD semantics to improve the OOD detection models for vision data based on \textbf{VLMs}.

\begin{figure}[h]
	\centering
	\subfigure[]{
		\label{fig.a}
		\includegraphics[width=0.9\columnwidth]{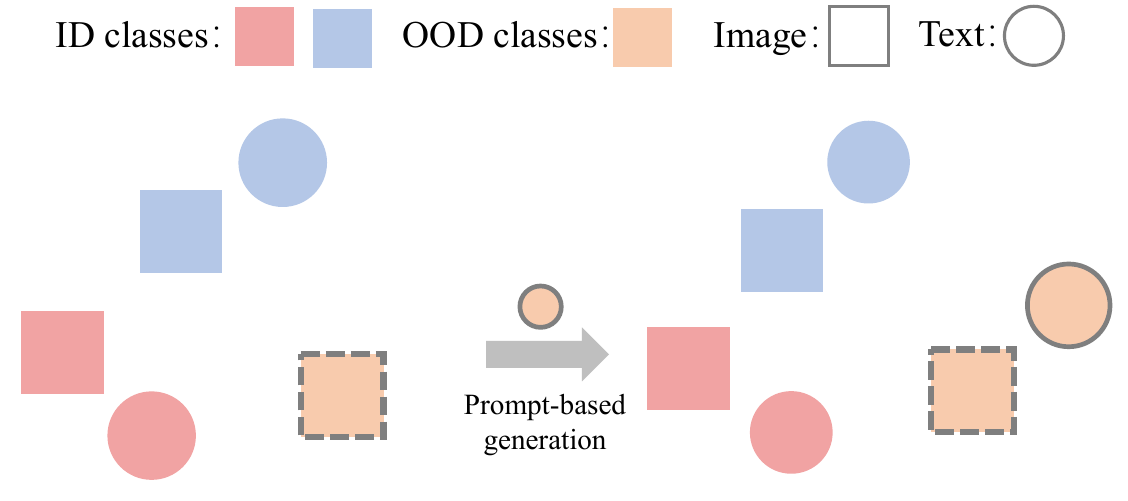}}\\
	\subfigure[]{
		\label{fig:gptgenerate}
		\includegraphics[width=0.9\columnwidth]{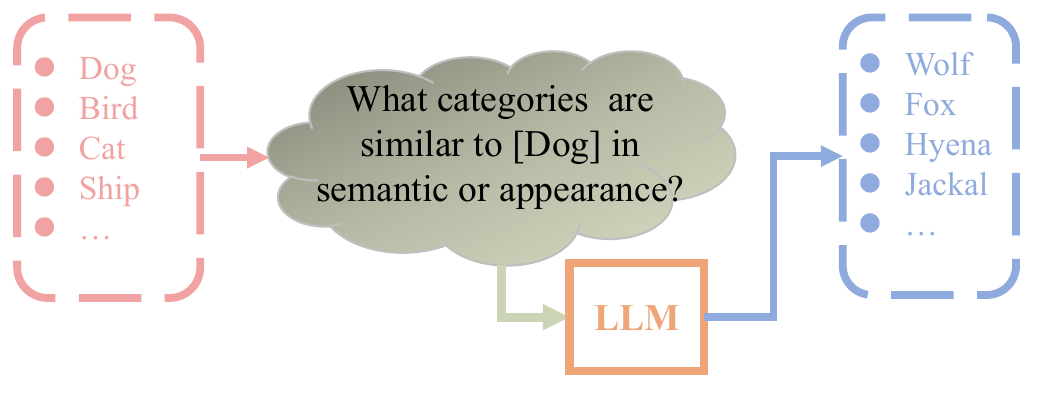}
	}
	\caption{In subfigure (a), the two modalities of the ID categories exhibit semantic similarity, while the orange rectangle indicates the unavailable of the OOD image modality, the orange circle signifies the available of the OOD text modality. The subfigure (b) illustrates how LLM generates peer-class labels based on ID class.}
	\label{fig:motivation}
\end{figure}



In real-world applications, models often lack access to OOD image information before deployment, while textual modality information is accessible. The question therefore arises: \emph{How do we define and generate auxiliary textual modality information that serves as OOD signals?} Images from two classes may have very similar visual characteristics but different properties in the semantic space. Based on the above observations, we propose to use the \textbf{LLM} to design prompts for generating OOD peer-classes that have semantic similarities with the same ID class (see figure \ref{fig:gptgenerate}). These classes contribute complementary cross-modal OOD information for the image modality. Additionally, within the accessible image modality, we use the Mixup strategy to construct OOD samples to extend the intra-modal OOD information. By establishing the Peer-Class Contrastive loss and Cross-Entropy loss for these two categories of samples and equipping the existing OOD detection method with the CLIP model, we present a framework for effective OOD detection. Our main contributions are summarized as follows:


\begin{itemize}[itemsep=2pt,topsep=2pt,parsep=1pt]
	\item We present to design appropriate prompts to generate OOD peer-class labels by LLM as an auxiliary modality to learn ID features with the compact decision boundary, for OOD detection.
	\item A novel loss, called PCCLoss, has been developed to effectively utilise the OOD information of textual modality to guide the learning of visual ID data.
	\item The results of experiment show that our proposed method achieves state-of-the-art performance on most benchmark datasets.
\end{itemize}

\section{Related Work}
\subsection{Out-of-distribution Detection with Single-modal}
OOD detection has emerged as a vital field in pattern recognition due to its importance for real-world applications. Traditional recognition systems assume a closed-set in which all test samples belong to the classes seen during training. In practise, however, it can happen that new or unseen classes exist during the test, which leads to the ood problem. OpenMax \cite{7780542} is an early method designed to address this problem. The core idea is to adjust the model's output probability distribution by employing open distribution calibration. DOC \cite{shu-etal-2017-doc} includes a classification loss function, combining cross-entropy with a penalty term for low confidence scores on OOD examples. OSRCI \cite{Neal_2018_ECCV} is to calibrate model outputs by establishing confidence intervals to better differentiate between unknown and known class samples. C2AE \cite{8953248} is enabling accurate recognition of known classes and rejecting unknown or novel classes by leveraging the auto-encoder's latent space and thresholding mechanism. CAC \cite{9423243} employs a decision thresholding mechanism to classify test samples. The distance between a test sample and each class anchor is computed and if the distance exceeds a predefined threshold. GFROR \cite{9157020} employs the discriminative classifier to make predictions based on the fused feature and discriminative feature with the GFD (Generative Feature Discrimination) module. G-ODIN \cite{9156473} involves temperature scaling and input perturbation during calibration to enhance the model's sensitivity to OOD samples. In the CSI \cite{DBLP:conf/nips/TackMJS20} method, contrastive learning with distributionally shifted instances is used to learn distinct and discriminative feature representations. Deep k-Nearest Neighbor(KNN) \cite{sun2022knnood} leverages feature embeddings extracted from a model and operates under the assumption that the test OOD samples are relatively far away from the ID data. Despite the good performance of these methods, they mainly focus on using information from single-modality for detection and fail to exploit more diverse cues from multi-modalities to improve performance.

\subsection{Out-of-distribution Detection with Multi-modal}
The self-attention mechanism of transformer models has revolutionized the field of natural language processing and inspired researchers to apply it to computer vision. Vision Transformer (ViT) \cite{dosovitskiy2021an} applied Transformer models to image understanding, and OpenAI’s CLIP extended ViT’s capabilities to enable the joint understanding of images and text. CLIP accomplishes this by leveraging the rich representations learned from both images and text, enabling unified understanding of multi-modal data.
%

Lately, there have been several approaches that have begun to incorporate multi-modal information, such as ZOC \cite{Esmaeilpour2021ZeroShotOD}. It uses the extended model to generate candidate unknown class names for each test sample and computes a confidence score based on both the known class names and candidate unknown class names for zero-shot OOD detection. A recent study \cite{fort2021exploring} performs zero-shot OOD detection through CLIP. It is assumed that a set of unseen labels originating from OOD datasets are provided as weak OOD information. In contrast, to make the OOD information in the unseen labels more effective, i.e., similar but different from the ID information, we propose to generate OOD peer-class labels with LLMs instead of the off-the-shelf unseen labels from other datasets.

\section{Method}
In this section, we present our method in detail and explain how to use multi-modal to help us in OOD detection based on pre-trained models. In our framework, the \textbf{VLM} primarily serves as a feature extractor, focusing on extracting features from images and text. \textbf{LLM} act primarily as auxiliary text generators and relate in particular to the generation of peer class labels based on ID classes.

\subsection{The Architecture}
\begin{figure*}[t]
	\centering
	\includegraphics[width=0.9\textwidth, height=0.4\textwidth]{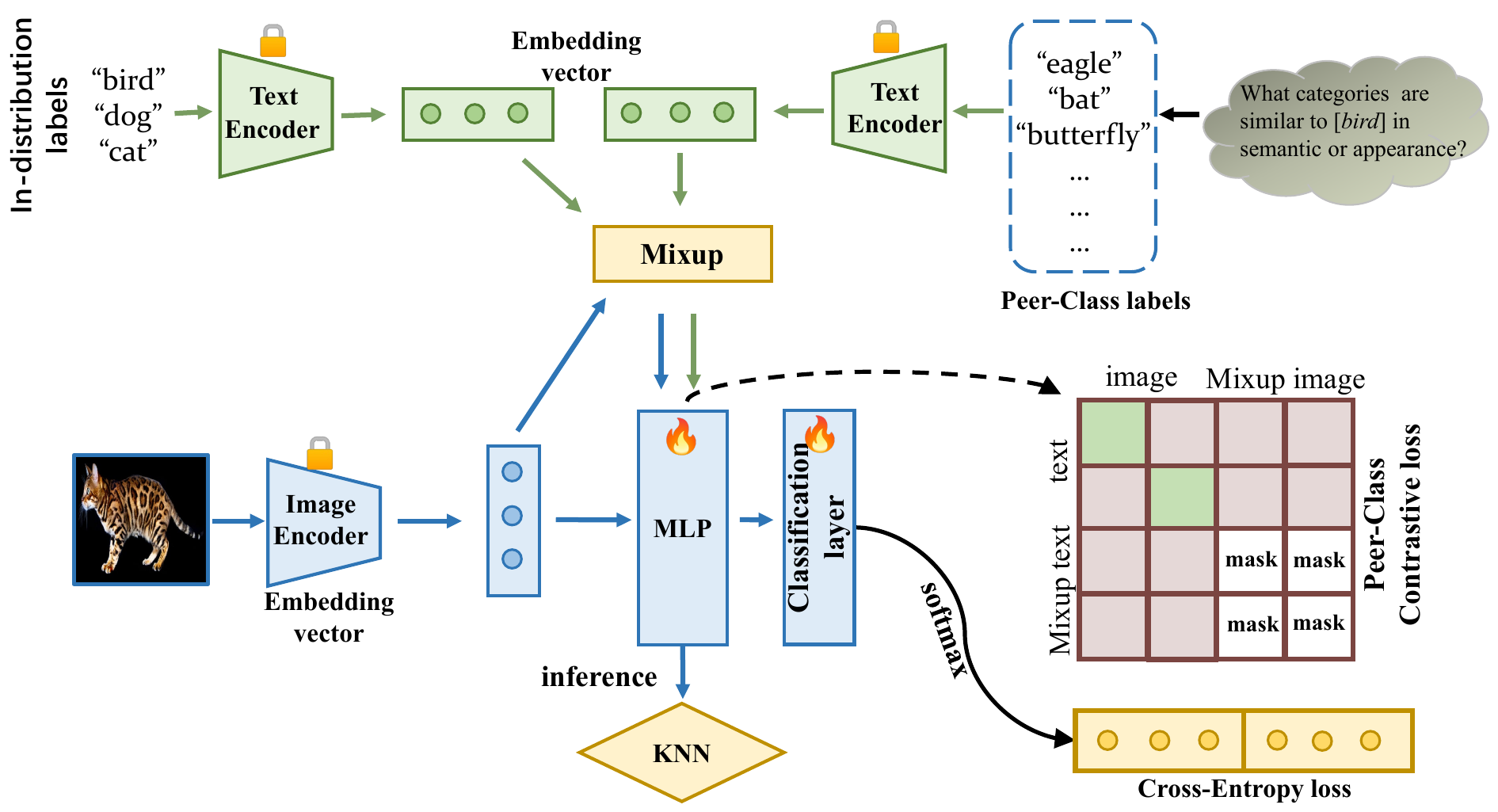}
	\caption{This diagram illustrates the framework of our approach. First, we use LLM to generate peer-class labels based on in-distribution (ID) classes. Then, these peer-class labels, along with the ID labels, are processed by the text encoder to extract text features, while the ID images are subject to feature extraction using the image encoder. Within the MLP, negative examples are generated by the mixup method and the Peer-Class Contrastive loss is used on each fully connected layer, the cross-entropy loss is added in the classification layer. Finally, we use KNN method to inference by using image features from MLP.}
	\label{fig3}
\end{figure*}

In our framework, we use CLIP as \textbf{VLM} as well as some baseline methods. In addition, we have extended CLIP with fully connected layers and the last classification layer, i.e., MLP. The dimension of the classification layer is determined based on the specific ID dataset and the number of generated peer-class labels (see figure \ref{fig3}).

\subsection{Peer-Class Generation by LLM}
The biggest challenge in OOD detection is that it is not possible to collect samples of OOD instances before deployment. However, if we focus on text modality, we can observe how easy it is to construct OOD samples in semantic space. GPT-3 \cite{NEURIPS2020_1457c0d6} is currently a highly prominent large-scale language model that is easily accessible on the Internet and exhibits strong text generation capabilities, which meets our requirements. We therefore instruct GPT-3 to generate the desired text information. For each class label $\mathcal{y}_i$  in the ID dataset, we formulate the prompt “what categories are similar to [class] in semantic or appearance” and instruct GPT-3 to generate the peer-class label for each class. We can select $n$ peer-class labels that are not present in the ID dataset for each class. Next, we combine the generated peer-class labels with the labels from the ID dataset and add textual descriptions like “This is a photo of a [CLASS]”.


\subsection{Inter-Modal OOD Generation}
The supplementary Peer-Class labels may still be insufficient to capture the underlying distribution of OOD samples, so we construct additional OOD samples within each modality. We use Mixup method \cite{10173749} $\mathcal{u}_i = \lambda \cdot \mathcal{x}_i + (1-\lambda)\cdot \mathcal{x}_{q(i)}$ to construct negative sample pairs, where $\lambda = 0.5$. For the image modality, we randomly select other class images $I_i$ from the same batch to construct negative image sample $I_{m^{(i)}}$ using Mixup. Each class of image features $I_i$ is paired with the corresponding textual description feature $T_i$. For the text modality, we randomly select one peer-class label description features $T_{p^{(i)}}$ from the same class and use Mixup to combine it with its textual description features to construct negative text sample $T_{m^{(i)}}$. To incorporate these positive and negative pairs into the contrastive loss, we propose the Peer-Class Contrastive loss (PCCLoss). Which is applied to each fc layers:

\begin{equation}
	\mathcal{L}_{PCC}^{l} = -log\frac{1}{N} \sum_{i=1}^{N}\frac{e^{(\langle I_i,T_i\rangle/ \tau)}}{\sum_{k\neq i} e^{{(\langle I_i,\{I_{m^{(k)}},\hat{T}_k,T_{m^{(k)}}\}\rangle/\tau)}}}
\end{equation}

where $\tau$ is a temperature hyper-parameter, $I_i$ and $T_i$ are paired image and textual description features. $\langle I_i,T_i\rangle = f(I_i)^Tf(T_i)$, where $f(\cdot)$ is fully layer connection projection. $\hat{T}_k$ represents the non-paired textual description features. $\{I_{m^{(k)}},\hat{T}_k,T_{m^{(k)}}\}$ is a set of negative examples. $N$ is the batch-size and $l$ represents the $l$-th fully connected layer. When the features have passed through all fully connected layers and reach the final classification layer. Our goal is to preserve the original classification capability of the features. Therefore, we add the Cross-Entropy loss in the last classification layer.

\begin{equation}
	\mathcal{L}_{CE} = - \frac{1}{N} \sum_{i=1}^{N} \sum_{c=1}^{M} \mathcal{y}_{ic}log(\mathcal{p}_{ic})
\end{equation}
where $\mathcal{y}_{ic}$ is the true label of the input sample and $\mathcal{p}_{ic}$ is the label predicted by the model.

\subsection{Training with Multi-Modal}

During the training phase, the images are processed by the CLIP image encoder, and the textual descriptions of the ID labels and the corresponding peer-class labels are passed through the text encoder to extract features. The image features pass through the subsequent fully connected layers, and in each layer, the text features are incorporated to construct a contrastive loss for learning.

\subsection{KNN for inference}
The KNN method calculate the k-th distance between the embedding of each test image and the training set and uses a simple threshold-based criterion to determine whether an input is an OOD sample or not. It is worth mentioning that the KNN method not only uses features from the penultimate layer, but also involves features from intermediate layers in the distance calculation. Therefore, in this paper, we calculate the distance using the KNN method\cite{sun2022knnood} for all features from the subsequently added fully connected layers.

\section{Experiments}
\subsection{Implementation and Settings}
\textbf{Implementation}: We use only the two encoders of CLIP: the image encoder and the text encoder, without changing or fine-tuning their parameters. We add three fully connected layers and one classification layer based on CLIP. Since the hidden state of the output of the image encoder is projected from 768 to 512 dimensions, we keep the dimension of the fully connected layers to match the output of the image encoder to ensure that the feature information is not lost. The dimension of the final classification layer is determined by the number of classes in the ID dataset and the number of corresponding peer-class labels. Before training, we start by querying GPT-3 with well-designed prompts which is “what categories are similar to [class] in semantic or appearance” to obtain peer-class labels for each ID class. We then select 3 peer-class labels from the generated labels for each class. The model were trained during 160 epochs with batch size 32, temperature $\tau=0.005$. We use SGD optimizer with a learning rate of 1e-5 and momentum of 0.99. We employ the stepLR schedule method with parameters step size of 30 and gamma of 0.25 for the optimizer. We use Faiss \cite{8733051}, a library for efficient nearest neighbor search, k of the KNN method is 200, threshold $\lambda$ is 0.95.

\begin{table*}[t]
	\centering
		\begin{tabular}{@{}lcccccc@{}}
			\toprule
			& CIFAR10  & CIFAR100 & CIFAR+10 & CIFAR+50 & TinyImageNet & \multicolumn{1}{c}{Average} \\ \hline
			\textbf{Original baselines}&          &          &          &          &              &                   \\ \hline
			OpenMax \cite{7780542}   & 69.5±4.4 & NR       & 81.7±NR  & 79.6±NR  & 57.6±NR      & 75.6    			\\
			DOC \cite{shu-etal-2017-doc}  & 66.5±6.0 & 50.1±0.6 & 46.1±1.7 & 53.6±0.5 & 50.2±0.5     & 58.2              \\
			G-OpenMax \cite{Ge2017GenerativeOF}  & 67.5±4.4 & NR       & 82.7±NR  & 81.9±NR  & 58.0±NR      & 75.9              \\
			OSRCI \cite{Neal_2018_ECCV}  & 69.9±3.8 & NR       & 83.8±NR  & 82.7±0.0 & 58.6±NR      & 77.2              \\
			C2AE \cite{8953248}      & 71.1±0.8 & NR       & 81.0±0.5 & 80.3±0.0 & 58.1±1.9     & 75.9              \\
			GFROR \cite{9157020}     & 80.7±3.0 & NR       & 92.8±0.2 & 92.6±0.0 & 60.8±1.7     & 84.0                        \\
			CSI \cite{DBLP:conf/nips/TackMJS20}& 87.0±4.0 & 80.4±1.0 & 94.0±1.5 & 97.0±0.0 & 76.9±1.2     & 87.0                        \\
			CAC \cite{9423243}       & 80.1±3.0 & 76.1±0.7 & 87.7±1.2 & 87.0±0.0 & 76.0±1.5     & 84.9              \\ \hline
			\textbf{Baselines with CLIP backbone} &          &          &          &          &              &               \\ \hline
			CLIP+CAC \cite{9423243}  & 89.3±2.0 & \textbf{83.5±1.2} & 96.5±0.5 & 95.8±0.0 & 84.6±1.7     & 89.9                        \\
			CLIP+G-ODIN \cite{9156473} & 63.4±3.5 & 79.9±2.3 & 45.8±1.9 & 92.4±0.0 & 67.0±7.1     & 69.8                        \\
			CLIP+MSP \cite{DBLP:conf/iclr/HendrycksG17}  & 88.0±3.3 & 78.1±3.1 & 94.9±0.8 & 95.0±0.0 & 80.4±2.5     & 87.3                        \\
			ZOC \cite{Esmaeilpour2021ZeroShotOD}  & 93.0±1.7 & 82.1±2.1 & 97.8±0.6 & 97.6±0.0 & 84.6±1.0     & 91.0               \\ \hline
			ours                     & \textbf{95.1±1.1} & 80.2±1.5 & \textbf{98.3±0.5} & \textbf{98.2±0.0} & \textbf{88.8±1.1} &\textbf{92.1}       \\  \bottomrule        
		\end{tabular}
		\caption{AUROC for the baselines and our proposed method. Best performance are bolded}
		\label{tab1}
\end{table*}

\textbf{Settings}: OOD detection problems can be categorized into near-OOD and far-OOD based on different datasets. This categorization is typically based on whether there exists visual or semantic similarity between the ID dataset and the OOD dataset. In this paper, we mainly focus on addressing the near-OOD problem. The difficulty level of an OOD detection task is commonly measured by the openness \cite{6365193} which is defined as follows:

\begin{equation}
	O = 1 - \sqrt{\frac{2 \cdot N_{train}}{N_{test}+N_{target}}}
\end{equation}
where $N_{train}$ is the number of known classes, $N_{target}$ is the number of unknown classes at test and $N_{test}$ is the total number of known and unknown classes at test. We perform experiments on (i) CIFAR-10 \cite{krizhevsky2009learning}, We randomly split it into 6 known and 4 unknown classes, Openness is 13.39\%. (ii) CIFAR+10, 4 non-animal classes of CIFAR10 as known and 10 randomly sampled animal classes of CIFAR100 \cite{krizhevsky2009learning} as unknown classes, Openness is 33.33\%. (iii) CIFAR+50 4 non-animal classes of CIFAR10 as known and 50 randomly sampled animal classes of CIFAR100 as unknown classes, Openness is 62.86\%. (iv) CIFAR100, randomly select 20 classes as known and 80 remaining classes as unknown classes, Openness is 42.26\%. (v) TinyImagenet \cite{article}, which is a subset of the ImageNet dataset, randomly select 20 classes as known and 180 remaining classes as unknown, Openness is 57.35\%.
All random sampling operations are repeated 5 times to get the average score. We report the commonly-used Area Under the Receiver Operating Characteristic Curve (AUROC) as evaluation metric for detecting OOD score.

\textbf{Baselines}: We compare our method with 12 OOD detection approaches, including OpenMax \cite{7780542}, MSP \cite{DBLP:conf/iclr/HendrycksG17}, G-OpenMax \cite{Ge2017GenerativeOF}, DOC \cite{shu-etal-2017-doc}, OSRCI \cite{Neal_2018_ECCV}, C2AE \cite{8953248}, GFROR \cite{9157020}, CSI \cite{DBLP:conf/nips/TackMJS20}, G-ODIN \cite{9156473}, CAC \cite{9423243}, ZOC \cite{Esmaeilpour2021ZeroShotOD}. Since our proposed method is based on CLIP model, we replaced the backbones with the image encoder of CLIP for MSP (represented by CLIP+MSP ), CLIP+G-ODIN and CLIP+CAC.
	
\subsection{Results and Discussion}
\begin{figure}[t]
	\centering
	\subfigure[original clip features]{
		\label{fig.sub.1}
		\includegraphics[width=4.2cm,height=2.8cm]{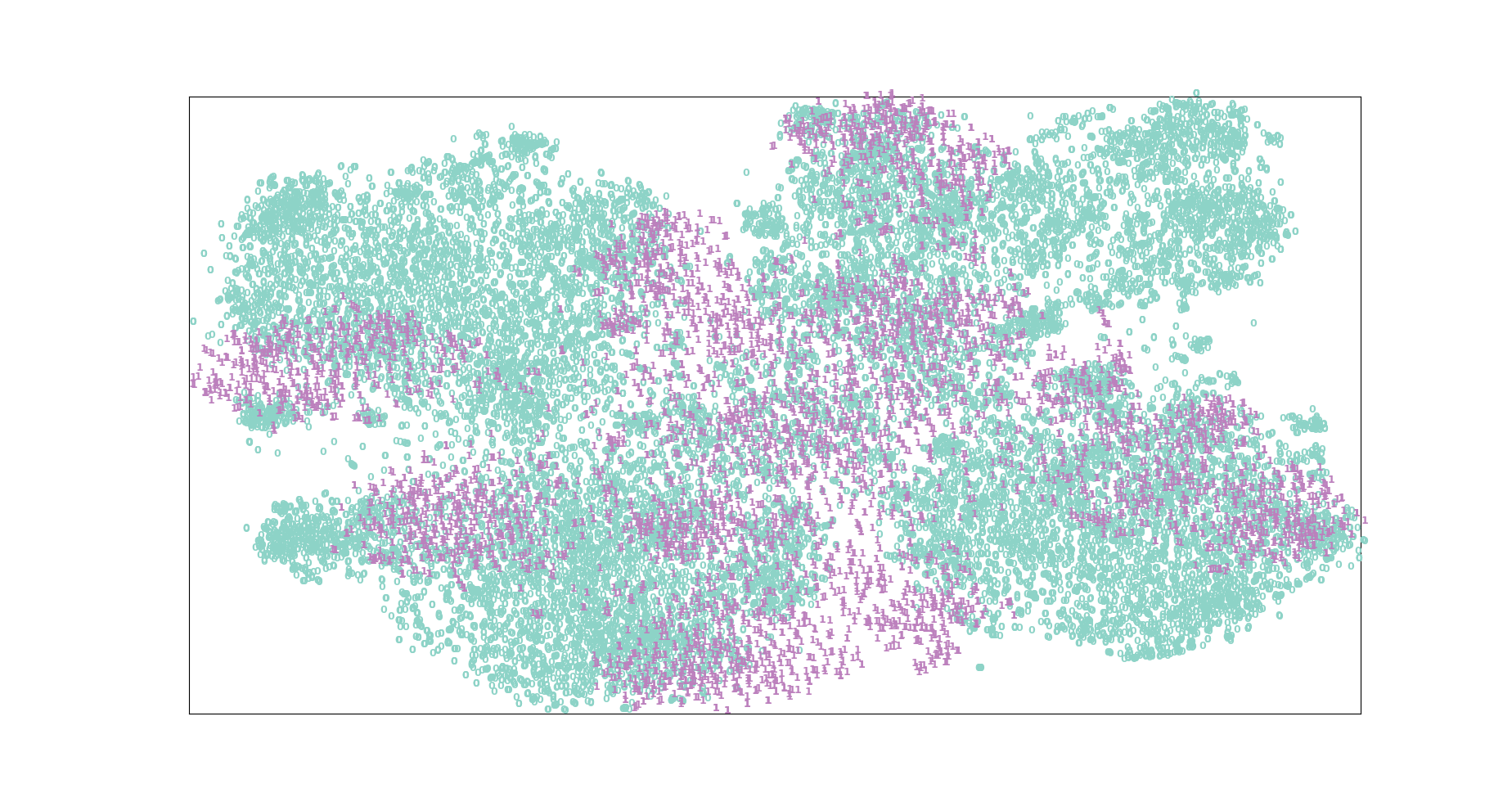}}\subfigure[features after training]{
		\label{fig.sub.2}
		\includegraphics[width=4.2cm,height=2.8cm]{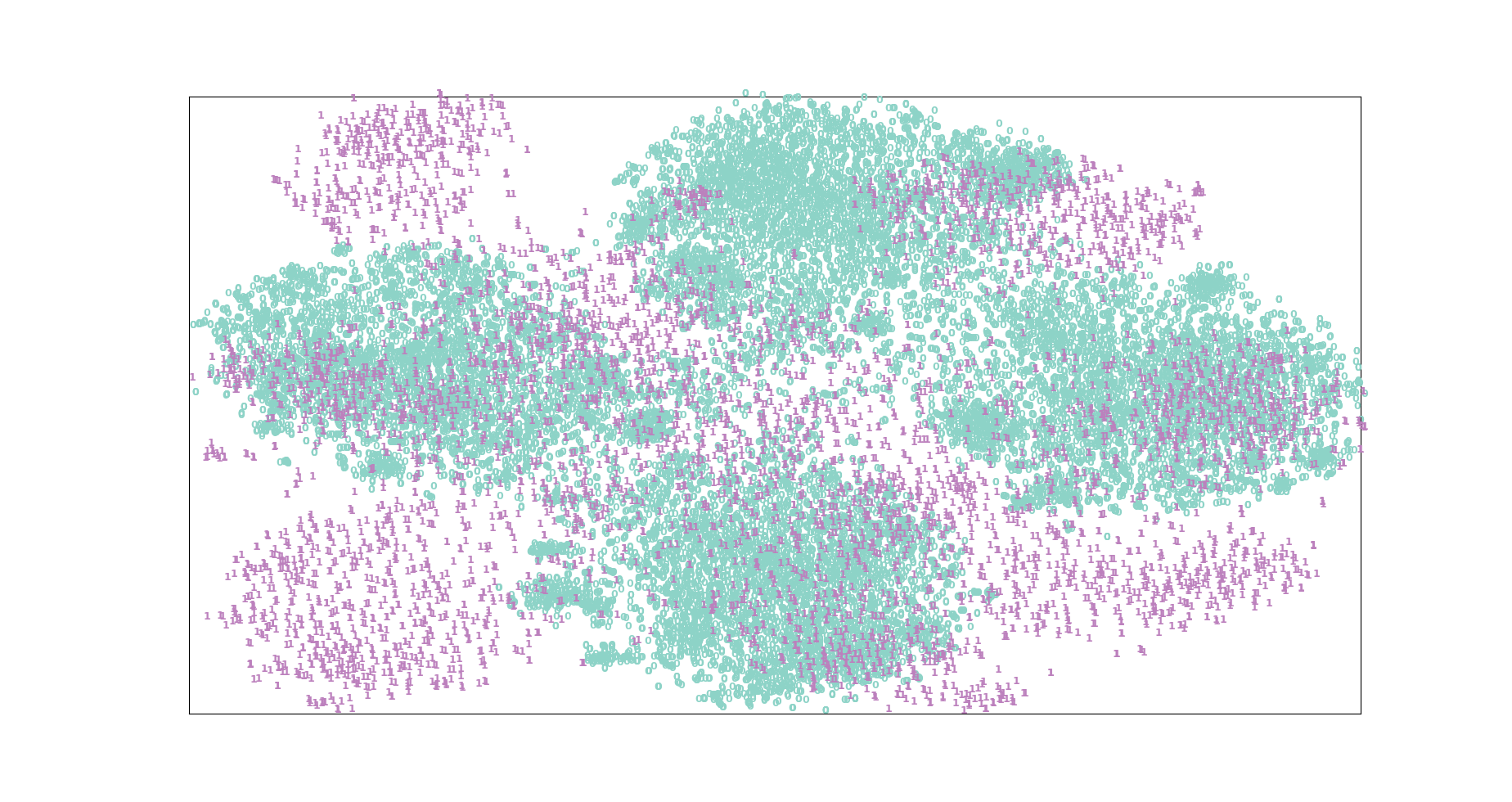}}
	\caption{The green dots means ID features and the pink dots means OOD features. After training, the features of the penultimate layer reveal that the characteristics of OOD have diverged from the clusters of the ID dataset.}
	\label{fig2}
\end{figure}

The results of these experiments are shown in Table \ref{tab1}. The experimental results show that our method outperforms all baselines. Considering that our trained model consists of only three fully connected layers and one classification layer without fine-tuning the CLIP model. Thus, the experimental results indicate that incorporating auxiliary peer-class textual modality can better assist in OOD detection. Our method is not highly dependent on the model, and other approaches can also potentially improve OOD detection performance by introducing text modality through the generation of peer-class information.
 (see Figure \ref{fig2})

\subsection{Ablation Study}
In this section, we will discuss the effectiveness of peer-class labels as an auxiliary text modality. We will test the two loss functions (PCCLoss and CELoss) separately and mixup method. Due to the use of all features from the fully connected layers during testing, if we only use CLIP features, there would be no training process and no intermediate fully connected layer features. To ensure a fair experiment, we will stack the CLIP features three times to achieve the same dimension. The experimental results were obtained on the TinyImageNet dataset (see Table \ref{tab2}). Since the Cross-Entropy loss is a typical classification loss, it may lead to a decrease in OOD detection performance when used alone. However, when combined with our proposed PCCLoss, it can increase the performance significantly.

\begin{table}[t]
	\centering
		\begin{tabular}{l|r}
			\toprule
			CLIP & 79.5  \\
			CLIP+CELoss & 75.9  \\
			CLIP+PCCLoss & 88.4 \\
			CLIP+CELoss+PCCLoss(without mixup) & 80.5\\
			CLIP+CELoss+PCCLoss & 88.8  \\
			\bottomrule
		\end{tabular}
		\caption{The results of ablation experiments show that incorporating peer-class labels and adding our proposed PCCLoss significantly improve the OOD detection performance.}
		\label{tab2}
\end{table}
	
\section{Conclusion}
In this work, we use the label information from the ID dataset to generate auxiliary text information to serve as OOD samples. During training, we use the mixup method to generate inter-modal OOD samples as negative pairs for constructing the Peer-Class Contrastive (PCC) loss. Existing single-modal methods can integrate our framework and try to improve the performance of OOD detection by using prompt learning to generate semantically
enriched text modality information.

\bibliographystyle{IEEEbib}
\bibliography{icme2023template.bib}

\end{document}